\documentclass{article}

\usepackage[T1]{fontenc}
\usepackage[preprint]{neurips_2026}

\usepackage{graphicx}
\usepackage{booktabs}
\usepackage{amsmath}
\usepackage{amssymb}
\usepackage{xcolor}
\usepackage{microtype}
\usepackage{caption}

\newcommand{\nObjects}{200}
\newcommand{\kSpeeds}{5}
\newcommand{\jitterSd}{0.25}
\newcommand{\OrSelf}{0.005}

\newcommand{\JitCorr}{0.96}
\newcommand{\BlSelf}{0.006}
\newcommand{\BlAnch}{0.165}
\newcommand{\BlErr}{0.368}
\newcommand{\BlAurocSelf}{0.70}
\newcommand{\BlAurocAnch}{1.00}
\newcommand{\BlCorr}{1.00}

\newcommand{\orMassR}{0.985}
\newcommand{\orFricR}{0.972}
\newcommand{\selFirstMiss}{7.1}

\newcommand{\selConsMiss}{8.6}

\newcommand{\selAnchMiss}{1.5}

\newcommand{\selPickedOracle}{98}
\newcommand{\selK}{4}
\newcommand{\noiseMax}{0.2}
\newcommand{\noiseMaxAurocBlind}{0.78}
\newcommand{\vjNOne}{500}
\newcommand{\vjNTwo}{300}
\newcommand{\vjQTrack}{0.03}
\newcommand{\vjQErr}{0.470}
\newcommand{\vjQSelf}{0.158}

\newcommand{\vjQCorrSelf}{0.10}
\newcommand{\vjQCorrAnch}{0.99}
\newcommand{\vjCTrack}{0.91}
\newcommand{\vjCErr}{0.337}
\newcommand{\vjCSelf}{0.159}
\newcommand{\vjCAnch}{0.195}
\newcommand{\vjCCorrSelf}{0.33}
\newcommand{\vjCCorrAnch}{0.92}
\newcommand{\vjSTrack}{0.05}
\newcommand{\vjSErr}{0.597}
\newcommand{\vjSSelf}{0.153}
\newcommand{\vjSAnch}{0.299}
\newcommand{\vjSCorrSelf}{0.12}
\newcommand{\vjSCorrAnch}{0.99}

\newcommand{\vjAnchOwnLtSwap}{73}
\newcommand{\vjSelfOwnLtSwap}{52}
\newcommand{\vjSelFirst}{0.645}
\newcommand{\vjSelCons}{0.392}
\newcommand{\vjSelSelf}{0.463}
\newcommand{\vjSelAnch}{0.306}
\newcommand{\vjSelOrac}{0.303}
\newcommand{\vjSelAnchOwn}{218}
\newcommand{\vjSelSelfOwn}{155}
\newcommand{\vjSelN}{300}
\newcommand{\vjSOneCTrack}{0.87}
\newcommand{\vjSOneSTrack}{0.04}
\newcommand{\vjSOneAnchOwnLtSwap}{75}
\newcommand{\vjSOneSelfOwnLtSwap}{46}
\newcommand{\vjSOneCCorrAnch}{0.96}
\newcommand{\vjSOneSCorrAnch}{0.99}

\title{IMPLY: Physically Anchored Consistency for World-Model Rollouts}

\author{%
  Aman Mehta\thanks{\texttt{heyamanmehta@gmail.com}} \\
  Independent Researcher
  \And
  Riya Baviskar\thanks{\texttt{riyakbaviskar@gmail.com}} \\
  Independent Researcher
}

\begin{document}
\maketitle


\begin{abstract}
A world model asked what happens if an object is pushed at several speeds produces several futures. If the model has the object in mind, those futures agree about it: each implies the same mass and friction. The consistency checks now used to vet world-action models ask whether a model's futures agree with each other, and none of them knows any physics. We show that this is not enough, and what to do instead. IMPLY reads the physics each rollout implies by inverting a simulator and scores a set of rollouts by how well one object explains all of them, anchored to two calibration pushes the model has observed. In a controlled setting, self-consistency gives a perfect score to a model that ignores the object and always predicts a typical push; anchoring exposes it (AUROC \BlAurocSelf{} versus \BlAurocAnch{}). On a real model, V-JEPA~2-AC adapted to the scene, the same thing happens. Given its own calibration pushes the model tracks the object (per-object correlation with the truth \vjCTrack{}); given another object's, it does not (\vjSTrack{}). Self-consistency cannot tell these apart, preferring the right evidence on \vjSelfOwnLtSwap{}\% of objects, chance level, while anchored disagreement prefers it on \vjAnchOwnLtSwap{}\% and correlates \vjCCorrAnch{}--\vjSCorrAnch{} with the rollouts' error. Used to choose among candidate rollout sets, it comes within $0.003$ of an oracle that sees the truth. A model that has internalised the wrong object is exactly as self-consistent as one that has internalised the right one; consistency has to be anchored to evidence.
\end{abstract}

\section{Introduction}

A robot with a world model plans by imagining: it proposes an action, generates the future that action would produce, and acts on the future it trusts. Whether a generated future can be trusted has therefore become the central reliability question for world-action models, and the field has converged on one kind of answer: consistency. A rollout is trusted if the actions it implies match the frames it shows, if several samples agree with one another, or if a forward and an inverse model agree about it. These signals are cheap, need no labels, and improve success rates when used to select among candidates. They also share a blind spot. They ask whether the model's outputs agree with each other, not whether they agree about the world. A model that always imagines the same plausible object, whatever object is actually on the table, is perfectly self-consistent and wrong.

We propose to measure consistency in physical terms. When a world model rolls out a push at several speeds, each rollout implies a mass and a friction for the pushed object, because a displacement at a known speed constrains both. If the model has the object in mind, the rollouts imply the same object. We recover the implied physics by inverting a simulator and score a set of rollouts by the residual of the single object that best explains all of them. The score uses no ground truth about the object. What it does use is evidence: two calibration pushes the model has observed, whose displacements anchor the fit. Without that anchor, we show, the score cannot distinguish a model that infers physics from one that recites a typical outcome.

We test the score twice. In a controlled setting (Section~\ref{sec:validation}) we build four stand-in models whose failures we choose: one with the true physics, one whose physics drifts between rollouts, one that ignores the object and predicts the typical push, and one in which nothing moves. Self-consistency catches the drifting model and the frozen one but scores the object-blind one exactly as well as the truth (AUROC \BlAurocSelf{}). Anchoring exposes it (\BlAurocAnch{}), tracks how wrong each model is, survives readout noise, and picks the true-physics candidate from a mixed set \selPickedOracle{}\% of the time. Then, on a learned model (Section~\ref{sec:models}), the same failure appears on its own. V-JEPA~2-AC~\citep{vjepa2}, adapted to the scene, tracks the object when given its own calibration pushes ($r=\vjCTrack{}$) and loses it when given another object's ($r=\vjSTrack{}$). Self-consistency cannot tell the two apart: it prefers the right evidence on \vjSelfOwnLtSwap{}\% of objects, chance level. Anchored disagreement prefers it on \vjAnchOwnLtSwap{}\%, correlates \vjCCorrAnch{}--\vjSCorrAnch{} with the rollouts' error, and selects among candidate rollout sets within $0.003$ of an oracle. The result reproduces across two adaptation seeds.

In short, we contribute (i) a consistency score for world-model rollouts stated in physical terms, the residual of the single object that explains every future, which needs no ground truth about the object; (ii) evidence, controlled and then on a learned model, that consistency without an anchor passes the failure that matters most, a model that ignores the object, and that two observed pushes are enough to catch it; and (iii) a demonstration that the anchored score predicts rollout error and chooses among futures nearly as well as an oracle that sees the truth.

\section{Related Work}

\textbf{Consistency signals for world-action models.} Action-state consistency ranks rollouts by whether the predicted actions produce the predicted frames and selects by consensus among futures~\citep{futurecompat}; it passes a rollout in which nothing moves, which the authors name background collapse. SC3-Eval enforces forward-inverse and cross-view consistency inside a video model used to evaluate policies~\citep{sc3eval}. Geometry-guided self-consistency clusters sampled action chunks and takes the medoid~\citep{ggsc}; MG-Select uses divergence from the model's own masked-input distribution~\citep{mgselect}; Pre-VLA trains a verifier~\citep{prevla}. None of these signals refers to the physical object being acted on. Ours does, and it needs the model to have observed the object interact.

\textbf{Hallucination in world models.} Hallucination concentrates where training coverage is thin and comes in perceptual, action-marginalised, and scene-diverging forms~\citep{halluc}. Our object-blind stand-in is action-marginalised hallucination in its purest form, and it is the case self-consistency cannot see.

\textbf{Identifying physics from interaction.} Inferring mass and friction from motion is classical~\citep{galileo, physics101}, and a short interaction history can identify a system in context~\citep{icwm}. CALIPER~\citep{caliper} showed that a linear readout on frozen video features infers latent physics from two calibration pushes and that swapping the calibration for another object's removes the gain. IMPLY turns the same calibration evidence into the anchor of a consistency test for generative rollouts.

\section{Implied Physics and Anchored Disagreement}
\label{sec:method}

\textbf{Setting.} An object of unknown mass $m$ and sliding friction $\mu$ rests on a table. A puck strikes it at a commanded speed $v$ and it slides a distance $d$. In the CALIPER environment~\citep{caliper} the map $f\colon (v, m, \mu) \mapsto \log d$ is deterministic given the object's geometry, and we tabulate it once on a grid ($19 \times 25 \times 21$ MuJoCo simulations with the canonical box) and interpolate. A world model is given the object's two calibration pushes at speeds $v^c_1, v^c_2$ with observed slides $d^c_1, d^c_2$, and is asked to roll out pushes at $K$ new speeds $v_1, \dots, v_K$. From each rollout we read a displacement $\hat d_k$.

\textbf{Implied physics.} A single $(v_k, \hat d_k)$ is consistent with a one-dimensional family of $(m, \mu)$; several at different speeds pin the pair down. We fit
\[
(\hat m, \hat \mu) = \arg\min_{m, \mu} \sum_{k} \big(f(v_k, m, \mu) - \log \hat d_k\big)^2
\]
by grid search over the table's range, which is cheap and immune to the multimodality that appears at the boundary of the range. The RMS residual of this fit is the \emph{self-disagreement} of the rollout set: zero when every rollout describes the same object, large when each describes a different one.

\textbf{Anchoring.} The \emph{anchored disagreement} includes the calibration observations in the same fit,
\[
\sum_{j=1}^{2} \big(f(v^c_j, m, \mu) - \log d^c_j\big)^2 + \sum_{k} \big(f(v_k, m, \mu) - \log \hat d_k\big)^2,
\]
so the rollouts are required to describe not just one object but \emph{this} object. The calibration slides are observed, not predicted, so the anchor costs nothing at test time beyond having watched the object be pushed.

\textbf{Reading displacement from a rollout.} For latent models we fit a ridge readout from the model's state to displacement on real clips, as in CALIPER; for pixel models we use frame differences, which in a fixed-camera scene read displacement to within a few millimetres. The validation below treats readout error as noise and asks how much of it the score tolerates.

\section{Controlled Validation}
\label{sec:validation}

Before applying the score to a learned model we ask what it does on models whose failure modes we control. Each stand-in produces $K=\kSpeeds{}$ rollouts per object for \nObjects{} CALIPER objects:

\begin{itemize}\itemsep1pt
\item \emph{Oracle}: the simulator with the object's true physics and geometry. Any useful score should give this near zero.
\item \emph{Jitter}: the simulator, but each rollout draws its own $(m, \mu)$ perturbed log-normally (sd \jitterSd{}) around the truth. A model with no stable world in its head.
\item \emph{Object-blind}: predicts the dataset-typical displacement for the speed and ignores the object. The pattern-matcher.
\item \emph{Collapse}: the object never moves. The background-collapse failure of~\citet{futurecompat}.
\end{itemize}

\begin{table}[t]
\centering
\caption{Stand-in models on \nObjects{} objects, $K=\kSpeeds{}$ speeds. Disagreement is the RMS residual in log displacement of the joint $(m,\mu)$ fit, without (self) and with (anchored) the two calibration observations. AUROC is the score's ability to separate a stand-in from the oracle. Self-consistency cannot see the object-blind model; anchoring can.}
\label{tab:synthetic}
\small
\begin{tabular}{lcccccc}
\toprule
& \multicolumn{2}{c}{Disagreement} & & \multicolumn{2}{c}{AUROC vs.\ oracle} & \\
\cmidrule(lr){2-3} \cmidrule(lr){5-6}
Stand-in model & self & anchored & log error & self & anchored & corr(anch., err.) \\
\midrule
Oracle (true physics) & 0.005 & 0.006 & 0.000 & -- & -- & -- \\
Jitter (no stable world) & 0.310 & 0.276 & 0.368 & 1.00 & 1.00 & 0.96 \\
Object-blind (typical slide) & 0.006 & 0.165 & 0.368 & 0.70 & 1.00 & 1.00 \\
Collapse (never moves) & 4.715 & 4.114 & 6.516 & 1.00 & 1.00 & 0.99 \\
\bottomrule
\end{tabular}
\end{table}

\begin{figure}[t]
\centering
\includegraphics[width=\linewidth]{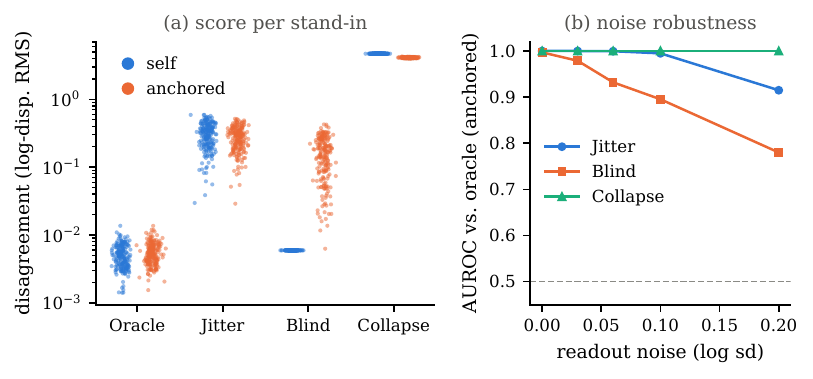}
\caption{Controlled validation on \nObjects{} objects. (a) Self and anchored disagreement per stand-in model; the object-blind model is invisible to self-consistency and exposed by anchoring. (b) Anchored AUROC against the oracle as readout noise grows. }
\label{fig:validation}
\end{figure}

\textbf{Self-consistency misses the pattern-matcher.} Table~\ref{tab:synthetic} and Figure~\ref{fig:validation}a show the result that motivates the method. The oracle's implied physics recovers the true object (mass $R^2$ \orMassR{}, friction \orFricR{}), so the inversion works, and its self-disagreement is \OrSelf{}. Jitter and collapse are separated from the oracle perfectly by self-disagreement alone. The object-blind model is not: its self-disagreement is \BlSelf{}, indistinguishable from the oracle, because it describes the same typical object every time, and one object explains its rollouts exactly. Its error against the truth (\BlErr{} in log displacement) is as large as jitter's. A consistency signal that only compares a model's outputs with each other cannot see this failure, and it is the failure that matters most in practice: a model that has learned what pushes usually do, and nothing about the object in front of it.

\textbf{Anchoring fixes it.} With the two calibration observations in the fit, the object-blind model's disagreement rises to \BlAnch{} and it separates from the oracle with AUROC \BlAurocAnch{}. Anchored disagreement correlates \JitCorr{} with rollout error within the jitter model and \BlCorr{} within the object-blind model, so it tracks not only whether a model is wrong but how wrong.

\begin{table}[t]
\centering
\caption{Top: anchored disagreement under multiplicative readout noise on every rollout's displacement; AUROC against the oracle at the same noise level. Bottom: choosing among \selK{} candidate rollout sets of mixed quality by anchored disagreement, versus taking the first sample or the consensus; miss is the distance between the chosen query-speed displacement and the truth.}
\label{tab:noise_sel}
\small
\begin{tabular}{lcccc}
\toprule
Readout noise (log sd) & oracle score & AUROC jitter & AUROC blind & AUROC collapse \\
\midrule
0.0 & 0.006 & 1.00 & 1.00 & 1.00 \\
0.03 & 0.021 & 1.00 & 0.98 & 1.00 \\
0.06 & 0.044 & 1.00 & 0.93 & 1.00 \\
0.1 & 0.071 & 0.99 & 0.90 & 1.00 \\
0.2 & 0.147 & 0.91 & 0.78 & 1.00 \\
\bottomrule
\end{tabular}\\[6pt]
\begin{tabular}{lcc}
\toprule
Selection rule & median miss (mm) & $\leq$10\,mm \\
\midrule
First candidate & 7.1 & 52\% \\
Consensus (median) & 8.6 & 52\% \\
Lowest anchored disagreement & 1.5 & 98\% \\
Oracle candidate (upper bound) & 1.6 & 98\% \\
\bottomrule
\end{tabular}
\end{table}

\textbf{Robustness to readout noise.} Displacements read from generated frames are not exact. Table~\ref{tab:noise_sel} (top) adds log-normal noise to every rollout's displacement before scoring. The oracle's own score rises with the noise, as it must, but the separation from the inconsistent stand-ins survives: at noise sd \noiseMax{}, several times the error of our frame-based readout, the object-blind model is still separated with AUROC \noiseMaxAurocBlind{}.

\textbf{Selection.} The practical use of a consistency score is to choose among candidates. For each object we draw \selK{} candidate rollout sets, one from the oracle and the rest from the stand-ins at random, add readout noise, shuffle, and select. Anchored disagreement picks the oracle candidate \selPickedOracle{}\% of the time and brings the median miss at the query speed to \selAnchMiss{}\,mm, against \selFirstMiss{}\,mm for the first sample and \selConsMiss{}\,mm for the consensus of the candidates (Table~\ref{tab:noise_sel}, bottom). Consensus does no better than the first sample here because the wrong candidates do not agree with each other either; what identifies the right one is agreement with the object's observed behaviour.

\section{A Learned World Model: V-JEPA 2-AC}
\label{sec:models}

\textbf{Setup.} We adapt V-JEPA 2-AC~\citep{vjepa2} to the CALIPER scene: the ViT-g encoder stays frozen and the 300M action-conditioned predictor is fine-tuned with its own latent-prediction objective on 1,500 training episodes (4,500 clips, 8 frames each), the puck's commanded speed standing in for the end-effector action. A ridge readout from the encoder's last-frame features to displacement, fit on real clips, reaches held-out $R^2 = 0.984$, so displacement can be read off a latent rollout. We train two variants that differ only in context: one sees the first two frames of the query push, the other additionally sees three frames from each of the object's two calibration pushes. Rollouts are deterministic, so the K=5 futures per object come from the five speeds. Truth is the simulator run at those speeds.

\begin{table}[t]
\centering
\caption{V-JEPA 2-AC rollouts on \vjNTwo{} held-out objects (\vjNOne{} for the model trained without calibration context). \emph{Tracks object}: correlation across objects between the mean predicted and mean true log slide. Self and anchored disagreement as in Section~\ref{sec:method}; the last two columns correlate each score with the rollouts' error against the simulator.}
\label{tab:vjepa}
\small
\begin{tabular}{lcccccc}
\toprule
& tracks object & log error & \multicolumn{2}{c}{disagreement} & \multicolumn{2}{c}{corr.\ with error} \\
\cmidrule(lr){4-5} \cmidrule(lr){6-7}
Context & $r$ & & self & anchored & self & anchored \\
\midrule
\multicolumn{7}{l}{\emph{Trained without calibration context}} \\
query start only & 0.03 & 0.470 & 0.158 & 0.244 & 0.10 & 0.99 \\
\multicolumn{7}{l}{\emph{Trained with calibration context}} \\
own calibration $+$ query start & 0.91 & 0.337 & 0.159 & 0.195 & 0.33 & 0.92 \\
other object's calibration $+$ query start & 0.05 & 0.597 & 0.153 & 0.299 & 0.12 & 0.99 \\
query start only & -0.00 & 1.136 & 0.472 & 0.612 & 0.02 & 1.00 \\
\bottomrule
\end{tabular}
\end{table}

\begin{figure}[t]
\centering
\includegraphics[width=\linewidth]{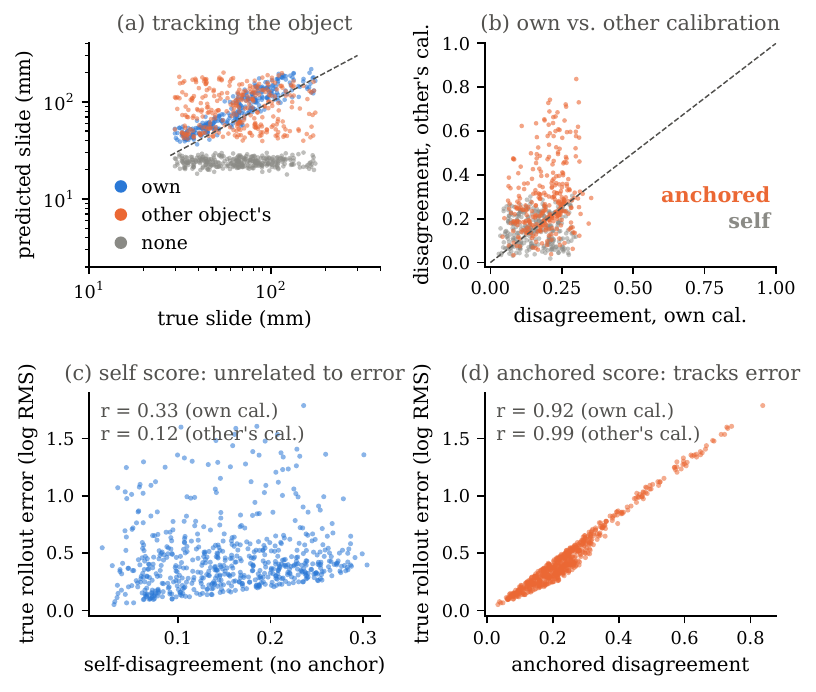}
\caption{V-JEPA 2-AC on \vjNTwo{} held-out objects. (a) Mean predicted against mean true slide per object, with the object's own calibration pushes in context, another object's, or none. (b) Each object's disagreement with its own calibration against its disagreement with another object's; points below the diagonal prefer the right evidence. Self-disagreement straddles the diagonal, anchored does not. (c, d) Does the score predict how wrong the rollouts are? Each point is one object: the score computed from the rollouts alone (x) against their true error measured with the simulator (y), both calibration conditions pooled. Self-disagreement says nothing about the error; anchored disagreement tracks it, so it can be used to decide which rollouts to trust without ground truth.}
\label{fig:vjepa}
\end{figure}

\textbf{Without evidence the model is object-blind, and only the anchored score sees it.} Given only the query's first frames, the model's futures rise with speed and land at a plausible scale, but they do not depend on which object is on the table: the per-object prediction correlates \vjQTrack{} with the per-object truth (Table~\ref{tab:vjepa}, Figure~\ref{fig:vjepa}a). This is the object-blind stand-in of Section~\ref{sec:validation}, produced by a real model. Its self-disagreement is low (\vjQSelf{}) and uncorrelated with error ($r=\vjQCorrSelf{}$); the anchored score correlates \vjQCorrAnch{} with error.

\textbf{With evidence the model infers the object, and the swap proves it.} Trained and prompted with the object's calibration pushes in context, the same predictor tracks the object ($r=\vjCTrack{}$) and its error falls from \vjQErr{} to \vjCErr{}. Replace the calibration with another object's and tracking collapses ($r=\vjSTrack{}$, error \vjSErr{}). The model is using the evidence. Self-consistency cannot tell the two cases apart (Figure~\ref{fig:vjepa}b): self-disagreement is \vjCSelf{} with the right evidence and \vjSSelf{} with the wrong evidence, and it is lower for the right evidence on only \vjSelfOwnLtSwap{}\% of objects, chance level. Anchored disagreement separates them (\vjCAnch{} versus \vjSAnch{}; lower for the right evidence on \vjAnchOwnLtSwap{}\% of objects) and tracks error in both conditions ($r=\vjCCorrAnch{}$ and \vjSCorrAnch{}), where self-disagreement does not (\vjCCorrSelf{}, \vjSCorrSelf{}). A model that has internalised the wrong object is exactly as self-consistent as one that has internalised the right one. Repeating the adaptation with a second data-order seed reproduces this (tracking \vjSOneCTrack{} with the right evidence, \vjSOneSTrack{} with the wrong; anchored lower for the right evidence on \vjSOneAnchOwnLtSwap{}\% of objects, self on \vjSOneSelfOwnLtSwap{}\%; anchored correlation with error \vjSOneCCorrAnch{} and \vjSOneSCorrAnch{}). Used to choose among the three candidate rollout sets per object (own calibration, another object's, none), anchored disagreement picks the own-calibration set for \vjSelAnchOwn{} of \vjSelN{} objects and brings the mean log error to \vjSelAnch{}, against \vjSelOrac{} for an oracle that sees the truth, \vjSelSelf{} for self-disagreement (which picks the right set for \vjSelSelfOwn{}), \vjSelCons{} for consensus, and \vjSelFirst{} for a random candidate.

\textbf{What this does and does not show.} The adaptation, not the pretrained model, is what learned to use calibration; the finding is about the score, which needed no ground truth to expose the object-blind regime, the wrong-evidence regime, and the error within each. The context-free variant's failure is by construction (it cannot know the object) and is reported because it is the regime most consistency checks are run in. We also adapted Cosmos-Predict2-2B (action-conditioned Video2World, LoRA rank 16, 4,000 iterations on the same clips). Its generated pushes render the scene faithfully but the box's displacement, read with a frame-difference readout that reaches $R^2=0.987$ on real clips, is the same at every commanded speed (26\,mm against a true range of 26--159\,mm) and changes only when the action is scaled far outside the training range. The adaptation did not produce action-dependent dynamics, so we do not score it; a full fine-tune or a larger model is the next step, and a pixel-space model remains the natural second test of the score.

\section{Limitations}

The inversion assumes a known simulator and a single contact type; for real scenes the table would come from a calibrated simulator or a learned forward model, and the score inherits its error. Anchoring requires that the model has observed the object being pushed, the same requirement CALIPER makes and one a manipulation system can meet by probing. The learned-model evidence comes from one architecture, and its use of calibration was learned during our adaptation rather than present in the pretrained checkpoint; what we establish is that the score detects both the presence and the absence of that use without ground truth, not that pretrained world models in general infer physics. Our attempt at a pixel-space model (Cosmos-Predict2, LoRA) did not produce action-dependent rollouts, so the pixel-readout path of the method is validated only on real and stand-in videos. Rollouts are deterministic here, so the candidate sets differ by context rather than by sampling; a stochastic model would let the score choose among samples directly. Finally, the stand-ins are caricatures of failure modes, useful for showing what a score can and cannot see, not a substitute for the learned-model study.

\bibliographystyle{plainnat}
\bibliography{references}

\end{document}